\documentclass[runningheads]{llncs}

\usepackage[english]{babel}

\usepackage{amsmath}
\usepackage{graphicx}
\usepackage[colorlinks=true, allcolors=blue]{hyperref}

\usepackage{mathptmx}
\usepackage{soul}\setuldepth{article}

\usepackage{amssymb}
\usepackage{makecell}
\usepackage{booktabs}
\usepackage{multirow}
\usepackage{array}
\usepackage{changebar}

\def\LAS{{\operatorname{\mathit{L-AS}}}}
\def\LAT{{\operatorname{\mathit{L-AT}}}}
\newcommand{\lasp}{\text{\it L-ASPIC }}

\newcommand{\af}{\text{\it AF }} 
\newcommand{\asp}{\text{ \it ASPIC$^+$ }}

\newcommand{\prem}{\text{ \tt Prem}} 
\newcommand{\conc}{\text{ \tt Conc}} 
\newcommand{\sub}{\text{ \tt Sub}} 
\newcommand{\dr}{\text{ \tt DefRules}} 
\newcommand{\tr}{\text{ \tt TopRule}} 
\newcommand{\rules}{\text{ \tt Rules}}
\newcommand{\lp}{\text{ \tt LastPrin}}
\newcommand{\lN}{\text{ \tt LastNorms}}

\begin{document}
\title{
An Argumentation-Based Legal Reasoning Approach for DL-Ontology
}
\titlerunning{An Argumentation-Based Legal Reasoning Approach for DL-Ontology}
\author{Zhe Yu\inst{1}
	\and
	Yiwei Lu\inst{2}
}
\authorrunning{Z. Yu and Y. Lu}
%
\institute{Institute of Logic and Cognition, Department of Philosophy, Sun Yat-sen University, Guangzhou 510275, China \\
	\email{yuzh28@mail.sysu.edu.cn}
	\and
Edinburgh Law School, University of Edinburgh, Edinburgh, UK
	\\\email{s2223086@ed.ac.uk}
}
\maketitle             

\begin{abstract}
Ontology is a popular method for knowledge representation in different domains, including the legal domain, and description logics (DL) is commonly used as its description language. 
To handle reasoning based on inconsistent DL-based legal ontologies, the current paper presents a structured argumentation framework particularly for reasoning in legal contexts on the basis of $ASPIC^+$, and translates the legal ontology into formulas and rules of an argumentation theory. 
With a particular focus on the design of autonomous vehicles from the perspective of legal AI, we show that using this combined theory of formal argumentation and DL-based legal ontology, acceptable assertions can be obtained based on inconsistent ontologies, and the traditional reasoning tasks of DL ontologies can also be accomplished. In addition, a formal definition of explanations for the result of reasoning is presented.
\end{abstract}

\section{Introduction}

Ontologies are popularly applied in knowledge representation, structuring and management. They can also be helpful in extracting and modeling information in the legal domain, as well as realizing the reasoning problem-solving based on legal ontologies. 
In the literature, many legal ontologies have been proposed, e.g. the Legal Knowledge Interchange Format (LKIF) core ontology builds on the Description Logics (DL) and LKIF rules \cite{Hoekstra07,Hoekstra09}; the Core Legal Ontology (CLO) based on the extension of the DOLCE (DOLCE+) foundational ontology \cite{Gangemi2005}; the LRI-Core ontology aimed at the legal domain grounded in common-sense \cite{Breuker2005}, and the Functional Ontology for Law (FOLaw) \cite{Valente95,Valente99}.

Law is based upon a dialectical process and reflecting preferences. It is expressed in natural language, which introduces ambiguity, interpretation problems and inconsistencies \cite{VanEngers2008}. 
For illustration, consider the following scenario,  where engineers are tasked with designing autonomous vehicles (AVs) that operates in a legal-compliant way.

\begin{example} \label{drunk ai}
 Currently, the law stipulates a number of behaviours that a human driver has to observe after accident has happened. This includes a duty to stay at the scene of an accident and to provide first aid if necessary and feasible. The design question now is, does a driverless car ``inherit'' this obligation in the same way it ``inherits'' from the human driver the duty to stop at a traffic light? A recent proposal by the Scottish and English Law Commissions differentiates between functions that are core to the safe operation of an AV and those that are auxiliary. When it hits somebody on the road, should it stop and do something, when sending information to police doesn't require stop and medical aid seems inapplicable? This depends on whether these functions are deemed ``core'' or ``auxiliary''.
 
As a complication, let's assume that there is one passenger in the car, but he is (illegally)  too drunk to do anything. In such a ``contrary to duty'' scenario, how should the AV car react when somebody is hit? Should it just report to the police and keep doing its original job (i.e., sending the passenger to destination as soon as possible)? If the injured party is likely to die without emergency medical attention, is it a legal requirement that the AV stops the only passenger from leaving and asks him to take the responsibility as a driver to give some help? Since the passenger is drunk, what if he does second harm to the injury or put himself in danger? 

\end{example}

We can easily note that, according to the current traffic law, for the scenario shown in the above example, the designer of AVs may be faced with different design options that lead to different choices, such as whether the intoxicated passenger should be asked to leave the car and help the injury.

However, the main description language of the current ontology is the Web Ontology Language (OWL) or OWL2, and the semantics are based on DL \cite{dlhandbook2003}.
It is difficult to perform reasoning based on DL under inconsistency, as well as to provide user-friendly explanations for how the answers of queries were obtained and why they were accepted.  
Since justification has been shown to be an effective type of explanation \cite{ye95}, meanwhile, formal argumentation \cite{dung95,dungkt09,GS04,MP13}, as a method for handling reasoning with inconsistent information, can provide various ways to explain why a claim is justified, therefore, in view of the above considerations, the current paper argues that argumentation is promising to provide an easy-to-explain method that can handle reasoning with inconsistent legal ontologies. 
The related basic idea was first presented in \cite{luyu20dl}. 

The rest of this paper is organized as follows. Section 2 first introduces the settings of \lasp framework, then translates the DL ontology into an argumentation theory, and presents a combined theory taking the design of a legal support system for AI vehicle based on Example \ref{drunk ai} as running examples. Section 3 shows how to perform inferences based on this approach, and illustrates our ideas with the examples. 
At last, section 4 concludes this paper.

\section{Argumentation Framework and Translation of DL}

\subsection{An Argumentation Framework for Legal Reasoning: \lasp}\label{sec-laspic}

We define the argumentation system particularly for legal reasoning based on \asp \\framework introduced in \cite{prakken10}, which is extended with a set of legal principles for argument preferences extraction, called \lasp.

\begin{definition}[Argumentation system]\label{def-las}
An \lasp argumentation system ($\LAS$) is a tuple $(\mathcal{L}, \mathcal{R}, n, \mathcal{P}, prin)$, where
	\begin{itemize}
		\item $\mathcal{L}$ is a set of formal language closed under negation ($\neg$);\footnote{We write $ \psi=-\varphi $ when $ \psi=\neg\varphi $ or $ \varphi=\neg \psi $ ($\varphi, \psi\in\mathcal{L}$).}
		\item $ \mathcal{R}=\mathcal{R}_{s}\cup \mathcal{N} $ is a set of strict inference rules ($ \mathcal{R}_{s} $) and legal norms ($ \mathcal{N} $) based on defeasible inference rules, 
		of the forms $ \varphi _{1},\ldots,\varphi_{n}\rightarrow \varphi $ and $ \varphi _{1},\ldots,\varphi_{n}\Rightarrow \varphi $ ($\varphi _{i},\varphi \in\mathcal{L}$), respectively; and $\mathcal{R}_{s}\cap \mathcal{N}=\emptyset $;
		\item $n$ is a naming function such that $n: \mathcal{N}\rightarrow\mathcal{L}$
		\item $\mathcal{P}$ is a set of legal principles; 
		\item $prin$ is a function from elements of $\mathcal{N}$ to elements of $\mathcal{P}$.
	\end{itemize}
\end{definition}

The standard \asp framework \cite{prakken10,MP13} defines the set of rules ($\mathcal{R}$) as consisting of two disjoint sets, i.e. a set of strict rules ($\mathcal{R}_s$) and a set defeasible rules. 
For normative reasoning in legal contexts, the argumentation system of \lasp replaces the set of defeasible rules with a set of legal norms ($\mathcal{N}$), and we assume that each legal norm is associated with a legal principle in $\mathcal{P}$ (whereas one legal principle may be associated with multiple norms), 
which is the primary legal principle on which the norm is based. In other words, for any legal norm $N\in\mathcal{N}$, $prin(N)\in\mathcal{P}$ is the basic legal principle upon which $N$ is constructed.

The reason we import the set $\mathcal{P}$ is twofold: 1) giving more legal semantics for reasoning results, making them more explainable; 2) supplying methods to solve complicated situations when different conflicting legal suggestions occur.

Given an $\LAS$, arguments can be constructed by rules starting from a set premises (knowledge base),  denoted as $\mathcal{K}$. And the tuple $(\LAS, \mathcal{K})$ is called an argumentation theory, denoted as $\LAT$. 
Usually, several basic parts of an argument can be noted: all the formulas in $\mathcal{K}$ that are used to build the argument (denoted as \prem), all its sub-arguments (denoted as \sub), all applied defeasible rules (denoted as \dr, while all applied rules can be denoted by \rules), the last rule it applied (denoted as \tr), and the consequent of the last rule (denoted as \conc). 

We define arguments constructed based on an $\LAT$ as follows.

\begin{definition}[Argument]\label{def-arg}
Let $\LAT=(\LAS, \mathcal{K})$ be an argumentation theory. An argument $\alpha$ constructed based on $\LAT^\varDelta$ has one of the following forms:
\begin{enumerate}
   \item $ \varphi $, if $ \varphi \in \mathcal{K}$, s.t.\footnote{such that} $\prem(\alpha)=\{\varphi\}$, $\conc(\alpha)=\varphi$, $\sub(\alpha)=\{\varphi\}$,  \\$\dr(\alpha)=\emptyset$, $\tr(\alpha)=undefined$;	
		\item $ \alpha_{1}$, $\ldots$, $\alpha_{n}$ $\rightarrow/\Rightarrow \psi$ if $ \alpha_{1} $, $ \ldots $, $\alpha_{n}$  are arguments, s.t.  there exists a strict/defeasible rule $ \conc(\alpha_{1}) $, $ \ldots $, $ \conc(\alpha_{n}) $ $\rightarrow/\Rightarrow \psi$ in $ \mathcal{R}$, and 
		$ \prem(\alpha)=\prem(\alpha_{1})\cup \ldots \cup \prem(\alpha_{n}) $, 
		$ \conc(\alpha)=\psi$, 
		$ \sub(\alpha)=\sub(\alpha_{1})\cup \ldots \cup \sub(\alpha_{n})\cup\{\alpha\}$, 
		\\$ \dr(\alpha)=\dr(\alpha_{1})\cup \ldots \cup \dr(\alpha_{n}) $ $(\cup\{  \conc(\alpha_{1}) $, $ \ldots $, \\$ \conc(\alpha_{n}) $ $\Rightarrow \psi)$, 
		$\tr(\alpha)=\conc(\alpha_{1}), \ldots, \conc(\alpha_{n})\rightarrow/\Rightarrow \psi$.
    
\end{enumerate}
\end{definition}

In addition, we define the notation for the (set of) last norms applied in an argument as follows.

\begin{definition}[Last Norms]\label{def-ln}
For any argument $\alpha$ constructed based on an \\$\LAT=(\LAS, \mathcal{K})$, let  $\texttt{LastNorms}(\alpha)=\emptyset$ if $\texttt{Norms}(\alpha)=\emptyset$, or \\$\texttt{LastNorms}(\alpha)=\{\texttt{Conc}(\alpha_1), \ldots, \texttt{Conc}(\alpha_n)\Rightarrow\psi$\} if $\alpha=\alpha_1, \ldots, \alpha_n\Rightarrow\psi$, otherwise \\$\texttt{LastNorms}(\alpha)=\texttt{LastNorms}(\alpha_1)\cup\ldots\cup \texttt{LastNorms}(\alpha_n)$.
\end{definition}

And we denote the set  $\{prin(N)|N\in\texttt{LastNorms}(\alpha)\}$ as $\texttt{LastPrin}(\alpha)$.

Given an $\LAT$, the inconsistency of legal information can be reflected by conflicts (attacks) among arguments. In general, an argument can be attacked on its uncertain parts or the consequent of uncertain elements. In \lasp, we assume that all the elements in $\mathcal{K}$ are uncertain and attackable. Then adapted from \cite{prakken10}, we define three types of attacks, i.e.: 1) \textit{rebutting} an argument on the consequent of norms, 2) \textit{undercutting} an argument on the norms it applies, and 3) \textit{undermining}  an argument on its premises. 

Formally, the attack relation is defined as follows.

\begin{definition}[Attacks]\label{def-att}
	Let $\alpha$, $\beta$, $\beta'$ be arguments constructed based on an $\LAT=(\LAS, \mathcal{K})$, $\alpha$ attacks $\beta$, iff $\alpha$ undercuts, rebuts or undermines $\beta$, where:
	\begin{itemize}
		\item $ \alpha $ undercuts $\beta $ on $ \beta' $, iff $ \beta' \in \sub(\beta)$ s.t.  $ \tr(\beta')=N\in \mathcal{N}$ and \\$\conc(\alpha)=-{n(N)}$\footnote{`$n(N)$' means that norm $N$ is applicable, therefore $\conc(\alpha)=-{n(N)}$ disables norm $N$.};
		\item $\alpha$ rebuts $\beta$ on $\beta' $, iff  $\beta' \in \sub(\beta)$ of the form $\beta''_1, \ldots, \beta''_n\Rightarrow\varphi$ and $ \conc(\alpha)=-\varphi $; 
		\item $ \alpha$ undermines $\beta $ on $\beta'$, iff $ \beta' =\varphi$ and $\varphi\in \prem(\beta)\cap\mathcal{K}$, s.t.  $\conc(\alpha)=-\varphi$.
	\end{itemize}
	
\end{definition}

When two arguments are in conflicts according to Definition \ref{def-att}, whether an attack can succeed as a defeat is determined by the preferences between arguments. In \lasp, preferences on the set of arguments are extracted according to the \textit{Democratic} approach for set comparison \cite{Cayrol92} and the \textit{last-link principle} \cite{MP13,prakken10} for the selection of elements being compared in terms of the priority ordering on the set of legal principles associated with norms. In the following definition, we call it a \textit{Democratic last-link principle}. 

Let $\mathcal{A}$ denote all the arguments constructed based on an $\LAT$, 
$\leqslant$ denote 
a preordering on $\mathcal{P}$ and $\trianglelefteq_{Dem}$ denote a set comparison based on the \textit{Democratic} approach \cite{Cayrol92}. The preference ordering $\preceq$ on $\mathcal{A}$ is defined as follows.

\begin{definition}[Argument ordering]\label{def-order}
Let $(\LAS, \mathcal{K})$ be an $\LAT$, for all $\alpha$, $\beta$ constructed based on it, 
$\beta\preceq \alpha$ under the Democratic last-link principle, iff \\
$\lp(\beta)\trianglelefteq_{Dem}\lp(\alpha)$, i.e.: 
\begin{enumerate}
    \item either $\lp(\alpha)=\emptyset$ and $\lp(\beta)\neq\emptyset$, or 
    \item $\forall P\in\lp(\beta)$, $\exists P'\in\lp(\alpha)$ s.t. $P\leqslant P'$. 
\end{enumerate}
\end{definition}

We write $\beta\prec\alpha$ iff $\beta\preceq\alpha$ and $\alpha\npreceq \beta$.

For the choice of comparative principles, the legal basis of the Democratic approach is that law in real life will protect preferred rights and benefits when it has to choose. For example, when two groups of legal rules are incompatible, the one containing the principle to protect human lives defeats the one aiming at protecting money (in many, but not all contexts). 
Meanwhile, the last-link principle is used primarily for legal applications and is more suitable for normative reasoning \cite{MPtutorial,prakken97}.

In order to get an output of acceptable conclusions, first we need to identify the justified arguments, which can be achieved by an argument evaluation process based on abstract argumentation frameworks (\textit{AF}) and argumentation semantics \cite{dung95}. Given an $\LAT$, an \af can be established based on the set of all the arguments, the attack relation between arguments (denoted as $att$) and argument ordering $\preceq$, defined as follows.

\begin{definition}[AF]\label{def-af}
Let $(\LAS, \mathcal{K})$ be an $\LAT$ and $SAF=\langle \mathcal{A}, att, \preceq\rangle$ a structured argumentation framework, where $\mathcal{A}$ is the set of all the constructed arguments, $att=\mathcal{A}\times\mathcal{A}$ is the set of attacks, and $\preceq$ is an ordering on $\mathcal{A}$. An abstract framework based on $\LAT$ is a tuple $AF=(\mathcal{A}, \mathcal{D})$, where $\mathcal{D}\subseteq att$ is a set of defeats, s.t. $\forall\alpha$, $\beta\in\mathcal{A}$, $(\alpha, \beta)\in\mathcal{D}$ iff: 1)$\alpha$ undercuts $\beta$, or 2) $\alpha$ rebuts/undermines $\beta$ and $\alpha\nprec\beta$.
\end{definition}

According to \cite{dung95}, we introduce the following basic argumentation semantics.

\begin{definition}[Argumentation Semantics]\label{def-semantics}
Let $(\mathcal{A}, \mathcal{D})$ be an \af. An extension $E\subseteq\mathcal{A}$ is \textit{conflict-free} iff $\nexists \alpha, \beta\in E$ s.t.  $(\alpha, \beta)\in\mathcal{D}$, and $\alpha$ is \textit{defended} by $E$ (or \textit{acceptable} w.r.t. $E$), iff $\forall{\beta}\in \mathcal{A}$, if $(\beta, \alpha)\in\mathcal{D}$, then $\exists\gamma\in {E}$ s.t. $(\gamma,  \beta)\in\mathcal{D}$. Then: 
	\begin{itemize}
		\item $E$ is an \textit{admissible set} iff  $E$ is  conflict-free and $\forall \alpha\in E$, $\alpha$ is defended by $E$; 
		\item $E$ is a \textit{complete} extension iff $E$ is admissible, and $\forall \alpha\in\mathcal{A}$ defended by $E$, $\alpha\in E$;
		\item $E$ is a \textit{grounded} extension iff $E$ is the minimal complete extension w.r.t. set-inclusion;
		\item $E$ is a \textit{preferred} extension iff $E$ is the maximal complete extension w.r.t. set-inclusion. 
	\end{itemize}
\end{definition}

Let $S\in\{co, gr, pr\}$ denote one of the above-mentioned argumentation semantics, $\mathcal{E}_S$ denote the set of all extensions obtained under $S$, and $E_S\in\mathcal{E}_S$ denote one of the extensions. An argument $\alpha$ is said to be \textit{accepted} w.r.t. $E_S$ if $\alpha\in E_S$. $\alpha$ is \textit{sceptically justified} under $S$ if $\forall E_S\in\mathcal{E}_S$, $\alpha\in E_S$, and $\alpha$ is \textit{credulously justified} under $S$ if $\exists E_S\in\mathcal{E}_S$ such that $\alpha\in E_S$. Then according to the accepted/justified arguments, we can identify the accepted conclusions.

\subsection{A Mapping from Description Logics to the Argumentation Theory}

The basic notions of description logic systems are \textit{concepts} and \textit{roles}, while complex concepts and roles can be constructed from basic ones by certain constructors. A description logic system contains two disjoint parts: the TBox and the ABox. TBox introduces the terminology, while ABox contains facts about individuals in the application domain. With different constructors, different description logics are constructed. In this paper, the legal ontology is mainly built upon the $\mathcal{ALC}$ expression \cite{ALC91,dlhandbook2003}, which is one of the basic language of description logics.

When inconsistency occurs in a DL ontology, traditional reasoners always issue an error message and stop reasoning. People have to find the partitions that cause the inconsistency and repair the problem. However, this `debugging' process could be very difficult, especially when the ontology is large. Considering the cost of computation and the difficulty of obtaining certain and complete information, this approach is inefficient or even impractical. Instead, in some cases, we may expect the reasoner can still perform reasoning based on the inconsistent knowledge and provide us a temporarily reasonable answer. Meanwhile, the users may need understandable explanations for the result. 
Based on \lasp, our idea is to perform such tasks by translating DL ontologies into argumentation theories. 

As specified in Definition \ref{def-las}, \lasp framework does not stipulate its logical language, only asks for a contradictory relation defined over it.\footnote{In paper \cite{prakken10,MP13}, it is a contrary relation, the current paper simplifies it for the time being.} Taking advantage of this unrestriction, in the current paper we adopt first order predicate logic, which is also the traditional logical basis of DL, to represent the elements in the set of language $\mathcal{L}$ of the combined argumentation theory $\LAT^\varDelta$. 

To translate the DL ontology into an argumentation theory, a \textbf{concept} $C$ as a unary predicate is represented as $C(x)$, while a \textbf{role} $P$ is represented as a binary predicate $P(x,y)$, where $x$, $y$ are individuals. $C\sqcap D$ and $C\sqcup D$ can be regarded as $C(x)\wedge D(x)$ and $C(x)\vee D(x)$ respectively, and the basic inference on concept expressions in DL is subsumption, written as $C\sqsubseteq D$. Formulas in the ABox are contained in the knowledge base $\mathcal{K}$ of $\LAT$. As for the TBox, inspired by \cite{simari10aai}, we translate the declarations in it into strict or defeasible rules, as shown in Table \ref{tab-mapping2}. In the table, $C$ and $D$ represent basic concepts, $P$ and $Q$ represent roles, $x$ and $y$ represent individuals, while $\alpha$ and $\beta$ represent free variables. Moreover, we use $\dashrightarrow$ to denote either $\rightarrow$ or $\Rightarrow$, which denote strict rules and defeasible rules respectively. A formula can only be translated into one type of rules according to the specified context.

\begin{table}
	\caption{Mapping from TBox of DL ontology to rules in the $\LAT$}\label{tab-mapping2}
	\centering
	\begin{tabular}{|l|l|}
		\hline
		TBox of DL &   rules in $\mathcal{R}$ of $\LAT$  \\
		\hline
		$C\sqsubseteq D$ & $C(x)\dashrightarrow D(x)$\\
        $C\equiv D$ & $D(x)\dashrightarrow C(x)$ \& $C(x)\dashrightarrow D(x)$\\
        $C\sqsubseteq \forall P.D$ & $C(x), P(x, y)\dashrightarrow D(y)$ \& $C(x)\dashrightarrow P(x, \alpha)$\\
        $P\sqsubseteq Q$&$P(x, y)\dashrightarrow Q(x, y)$\\
        $P\equiv Q$& $P(x, y)\dashrightarrow Q(x, y)$ \& $Q(x, y)\dashrightarrow P(x, y)$\\
        $C\sqsubseteq\exists P.D$& $C(x)\dashrightarrow P(x, \beta)$ \& $C(x), P(x, \beta)\dashrightarrow D(\beta)$\\
        $C\sqcap D\sqsubseteq \emptyset$&$C(x)\dashrightarrow \neg D(x)$ \& $D(x)\dashrightarrow \neg C(x)$\\
        $C\sqsubseteq D\sqcup Z$ & $C(x)\dashrightarrow D(x)\vee Z(x)$ \& $C(x), \neg D(x)\dashrightarrow Z(x)$ \& $C(x), \neg Z(x)\dashrightarrow D(x)$ \\
        $C\sqsubseteq D\sqcap Z$ & $C(x)\dashrightarrow D(x)$ \& $C(x)\dashrightarrow Z(x)$\\
       
		\hline
	\end{tabular}
\end{table}

Note that in Table \ref{tab-mapping2},  we translate $C\sqsubseteq D\sqcup Z$ in TBox into three rules. To clarify this translation, assume that there exists a declaration formed of $C\sqsubseteq D\sqcup Z$ in the DL ontology. According to our translation, three arguments $A_1$, $A_2$ and $A_3$ may be constructed, with $TopRule(A_1)=C(x)\dashrightarrow D(x)\vee Z(x)$,  $TopRule(A_2)=C(x), \neg D(x)\dashrightarrow Z(x)$ and $TopRule(A_3)=C(x), \neg Z(x)\dashrightarrow D(x)$. 
Meanwhile, since the rule $C(x), \neg D(x)\dashrightarrow Z(x)$ is applied, there must exist an argument $A_2'\in Sub(A_2)$ such that $Conc(A_2')=\neg D(x)$. For the same reason, there exists an argument $A_3'\in Sub(A_3)$ such that $Conc(A_3')=\neg Z(x)$. Apparently, it is counter-intuitive if $A_1$, $A_2'$ and $A_3'$ are accepted simultaneously because $\neg D(x)$ and $\neg Z(x)$ together contradict $D(x)\vee Z(x)$. However, based on Definition \ref{def-att}, $A_2'$ and $A_3'$ are conflicting with $A_3$ and $A_2$ respectively. As a consequence, if we have a declaration of the form $C\sqsubseteq D\sqcup Z$ in the TBox, $\{A_1, A_2', A_3'\}$ or $\{A_1, A_2, A_3\}$\footnote{The extensions of $ASPIC^+$ framework should be closed under sub-arguments. \cite{MP13}} will not be the subset of any extensions under classical argumentation semantics since conflict-freeness is a basic property. Therefore, with this translation, the output of an argumentation theory should be in line with human intuition. Meanwhile, it may reserve more expressiveness.\footnote{In paper \cite{simari10aai,grosof03},
when translate DL sentences into rules of logic programs and the $DeLP$ system, DL axioms that generate a rule with a disjunction in its head cannot be represented, so that lost some expressiveness.}

In a legal ontology, legally reasoning rules such as traffic rules and relevant policies will be allocated into the Tbox, while explicit legal designs such as an AI car should stop or not will be in the ABox. Consider the scenario in \textbf{Example} \ref{drunk ai} and assume that in an AV design, the driver should take the responsibility as current legal concept ``driver''. When an AV car hits somebody named ``Injury1'' on the road, it will ask the only passenger named ``PS1'' to leave the car and help the injured party, no matter whether he/she is sober. The information of explicit designs can be captured in the ABox and the TBox of a legal ontology describing the relevant assertions and rules as follows.

\begin{example}[DL encoding  of Example \ref{drunk ai}]
\label{exp-ontology}
\begin{equation*}
T=\left\{
\begin{array}{l}
Driver\sqsubseteq Sober;\ 
Sober\sqcap Intoxicated\sqsubseteq\emptyset;\ 
Intoxicated\sqcap LeaveCar\sqsubseteq \emptyset; \\
Driver\sqcap Intoxicated\sqsubseteq BeRevokedDrivingLicense\\\sqcap TakeCriminalResposibility; \\
\exists hitAndRun.Injury\sqsubseteq TakeCriminalResposibility; \\
\exists hitAndRun.causeDeath \sqsubseteq AggravatedPunishment; \\
\exists hitAndRun.Injury\sqcap Drive\sqcap Intoxicated \sqsubseteq AggravatingPunishment; \\
CauseAccident\sqcap Injury\sqsubseteq \exists transferToSafePlace.Injury; \\
CauseAccident\sqcap NeedEmergencyAid.Injury\sqsubseteq doNecessaryAid; \\
(transferToSafePlace\sqcup doNecessaryAid)\sqcap\neg LeaveCar\sqsubseteq\emptyset
\end{array}
\right\}
\end{equation*}

\begin{equation*} A=\left\{
  \begin{array}{l}
Driver(PS1);\ Intoxicated(PS1);\ hitAndRun(PS1, Injury1);\   Injury(Injury1); \\causeDeath(PS1, Injury1);\ CauseAccident(PS1);\
 NeedEmergencyAid(Injury1)
\end{array}
\right\}
\end{equation*}
\end{example}

According to the translation of ABox and TBox, we can represent DL ontology by argumentation theory based on \lasp.

\subsection{Combining the DL-based Legal Ontology with \lasp}

Basically, the idea is that the assertions in the ABox can be used as the premises of arguments in the argumentation theory, while according to the mapping that shown in Table \ref{tab-mapping2}, the terminologies in the TBox can be used as rules for establishing arguments. Accordingly, we use $T$ and $A$ to denote the sets of terminologies and assertions respectively.  
Let $\varDelta=(T, A)$ be a legal ontology for AV based on description logic, $(\LAS, \mathcal{K}^A)$ is an argumentation theory about $\varDelta$ (denoted as $\LAT^\varDelta$), where $\LAS=(\mathcal{L}, \mathcal{R}^T, n, \mathcal{P}, prin)$ such that $\mathcal{R}^T$ is the set of rules corresponding to $T$ based on Table \ref{tab-mapping2}, and $\mathcal{K}^A$ is the set of premises based on $A$.

An argumentation theory based on a DL-based legal ontology can be defined as follows.

\begin{definition}\label{def-DLarg}
Let $\varDelta=(T, A)$ be a legal ontology,  $\LAT^\varDelta=(\LAS, \mathcal{K}^A)$ is an argumentation theory about $\varDelta$, where $L$-$AS=(\mathcal{L}, \mathcal{R}^T, n, \mathcal{P}, prin)$ such that 
$\mathcal{R}^T$ is the set of rules corresponding to $T$ based on Table \ref{tab-mapping2}, and $\mathcal{K}^A$ is the set of premises based on $A$.
\end{definition}

In paper \cite{prakken10,MP13}, properties for a well-defined argumentation theory are specified. Likewise, a well defined $\LAT^\varDelta$ should also meet some requirements when transforming a legal ontology that contains inconsistencies into an combined argumentation theory. 
Let $Cl_{R_s^T}(Q)$ denote the closure of $Q\subseteq\mathcal{L}$ under strict rules, and $Q\vdash\varphi$ denote that there exists an argument $\alpha$ with $\rules(\alpha)\cap\mathcal{N}=\emptyset$, such that $\prem(\alpha)\subseteq Q$ and $\conc(\alpha)=\varphi$, and we propose the following definition for a well defined $\LAT^\varDelta$.

\begin{definition}\label{def-wd}
Let $(\mathcal{L}, \mathcal{K}^A, \mathcal{R}^T, n, \mathcal{P}, prin)$ be an $\LAT^\varDelta$. We say it is well defined, if $\forall\alpha$ constructed based on it, $\nexists\varphi$ such that $\prem(\alpha)\vdash\varphi,-\varphi$, and it is:
\begin{itemize}
    \item closed under contraposition or transposition, i.e. iff either: 
    \begin{enumerate}
        \item for all $Q\subseteq\mathcal{L}$ and $\varphi\in Q$, $\psi\in\mathcal{L}$, if $Q\vdash\psi$, then $Q\setminus\{\varphi\}\cup\{-\psi\}\vdash-\varphi$;
    \item or if $\varphi_1,\ldots, \varphi_n\rightarrow\psi\in\mathcal{R}_s^T$, then for each $i=1\ldots n$, there is \\
		$\varphi_1,\ldots, \varphi_{i-1},  -\psi, \varphi_{i+1}, \ldots \varphi_n\rightarrow-\varphi_i\in\mathcal{R}_s^T$;
    \end{enumerate}
    \item classical: iff for any minimal set $Q$ such that $\exists\varphi$, $Q\vdash\varphi, -\varphi$, it holds that $\forall\varphi\in Q$, $Q\setminus\{\varphi\}\vdash-\varphi$. 
\end{itemize}
\end{definition}

According to Definition \ref{def-wd}, when we find that two arguments constructed based on an $\LAT^\varDelta$ are conflicting with each other, at least one of them must contain uncertain elements.

\section{Inferences Based on the Argumentation Theory}

Given a legal ontology particularly for AVs design, we can construct an $\LAT^\varDelta$ based on the \lasp framework, as shown in \textbf{Example} \ref{exp-lesac}.

\begin{example}[A combined argumentation theory]\label{exp-lesac}
Given a legal ontology $\varDelta=(T, A)$ for AV, as shown in Example \ref{exp-ontology}. 
$\LAT^\varDelta=(\mathcal{L}, \mathcal{K}^A, \mathcal{R}^T, n, \mathcal{P}, prin)$ is an argumentation theory instantiated by $\varDelta$, such that

\begin{equation*}
\mathcal{N}
    =\left\{
    \begin{array}{l}
    r_1: Driver(x)\Rightarrow Sober(x);\\
    r_2: Intoxicated(x)\Rightarrow \neg LeaveCar(x); \\
    r_3: Driver(x), Intoxicated(x)\Rightarrow BeRevokedDrivingLicense(x); \\
    r_4: Driver(x), Intoxicated(x)\Rightarrow TakeCriminalResposibility(x);\\
    r_5: hitAndRun(x,y)\Rightarrow TakeCriminalResposibility(x);\\
    r_6: hitAndRun(x,y), causeDeath(x, y)\Rightarrow AggravatedPunishment(x);\\
    r_7: hitAndRun(x,y), Driver(x), Intoxicated(x)\Rightarrow AggravatedPunishment(x); \\
    r_8: CauseAccident(x), Injury(y) \Rightarrow transferToSafePlace(x, y);\\
    r_9: CauseAccident(x), Injury(y), NeedEmergencyAid(y) \Rightarrow doNecessaryAid(x, y) \\
    \end{array}
    \right\}
\end{equation*}
\begin{equation*}
    \mathcal{R}_{s}\footnote{Rules $r_{10}'$, $r_{11}'$ and $r_{12}'$ are the transposed rules of rule $r_{10}$, $r_{11}$ and $r_{12}$, respectively.
    }=\left\{
    \begin{array}{l}
   r_{10}: Sober(x)\rightarrow \neg Intoxicated(x);\\
    r_{10}': Intoxicated(x)\rightarrow \neg Sober(x);    \\
    r_{11}: transferToSafePlace(x, y) \rightarrow LeaveCar(x);\\
    r_{11}': \neg LeaveCar(x)\rightarrow \neg transferToSafePlace(x, y); \\
    r_{12}: doNecessaryAid(x, y) \rightarrow LeaveCar(x);\\
    r_{12}': \neg LeaveCar(x)\rightarrow \neg doNecessaryAid(x, y) \\
   
    \end{array}
    \right\}
\end{equation*}
\begin{equation*}
    \mathcal{K}^{A}=\left\{
    \begin{array}{l}
Driver(PS1); Intoxicated(PS1); hitAndRun(PS1, Injury1);   Injury(Injury1); \\causeDeath(PS1, Injury1); CauseAccident(PS1);
 NeedEmergencyAid(Injury1)
 \end{array}
 \right\}
\end{equation*}
\begin{equation*}
    \mathcal{P}=\left\{
    \begin{array}{l}
p_1: Human\ lives\ should\ be\ protected\ as\ a\ priority; \\
p_2: AI\ products\ should\ avoid\ extra\ risk\ about\ safety\ for\ their\ users;\\
p_3: People\ should\ avoid\ putting\ others\ into\ dangerous\ by\ his\ own\ behaviours,\ \\\qquad and\ should\ bear\ corresponding\ responsibility. \\
 \end{array}
 \right\}
\end{equation*}

$prin(r_1)=p_3 $; $prin(r_2)=p_2$;  $prin(r_3)=p_3$; $prin(r_4)=p_3$; $prin(r_5)=p_3$; 

$prin(r_6)=p_3$; $prin(r_7)=p_3$; $prin(r_8)=p_1$; $prin(r_9)=p_1$
\end{example}

\begin{figure}
	\includegraphics[width=\textwidth]{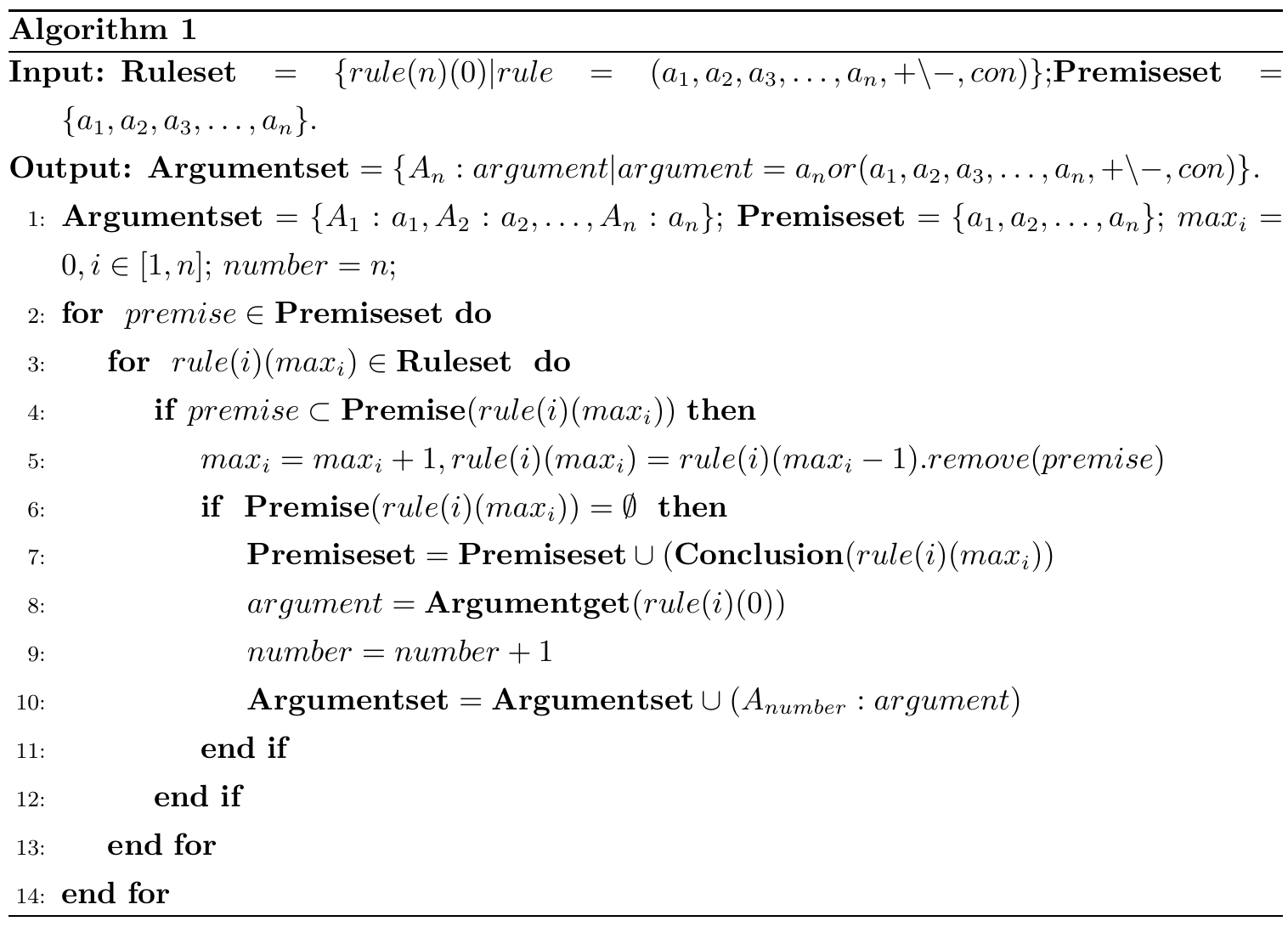}
	\caption{The procedure to get all the arguments of $AT$}\label{fig-code}
\end{figure}

To perform reasoning, one of the key points is to make sure all the arguments that can be constructed in the argumentation system are contained in the set $\mathcal{A}$.
In related works, how to achieve this task is often neglected without proper reason. The current paper presents an approach to fulfill this goal and the pseudocode is shown in Figure \ref{fig-code}, where $ a_1, a_2,\dots a_n$ denote premises, $+$ and $-$ denote the strict rules and the defeasible rules respectively, and $con$ denotes the conclusion, such that the rules and arguments have the same format $(a_1,a_2,a_3,\dots,a_n,+ \backslash -,con)$. Function $Premise(x)$ is used to obtain the premises of an argument $x$, $Conclusion(x)$ is used to obtain the conclusion of an argument $x$, and $Argumentget(r)$ is used to obtain an argument through rule $r$. The basic idea of this approach is to traverse the set of premises and set of rules respectively, and hide the premises/antecedents that appear in the body of rules. When all the antecedents of a rule are hidden, an argument will be constructed through the rule and its conclusion will be added to a set of premises/antecedents. This cyclic process continues until no new antecedent shows up. In this way, we can obtain a finite set of arguments containing all the available arguments according to the argumentation theory.

\begin{example}[Example \ref{exp-lesac} continued: arguments.]\label{exp-arg}

According to the $\LAT^\varDelta$ in Example \ref{exp-lesac}, we can get the following 22 arguments:

\begin{tabular*}{0.9\textwidth}{@{\extracolsep{\fill} }ll}
  $\alpha_1: Driver(PS1)$&
  $\alpha_2: Intoxicated(PS1)$\\
  $\alpha_3: Injury(Injury1)$&
  $\alpha_4: hitAndRun(PS1, Injury1)$  \\
  $\alpha_5: CauseAccident(PS1)$&
  $\alpha_6: causeDeath(PS1, Injury1)$\\
  $\alpha_7: NeedEmergencyAid(Injury1)$&
  $\alpha_8: \alpha_1\Rightarrow Sober(PS1)$\\
  $\alpha_{9}: \alpha_2\Rightarrow \neg LeaveCar(PS1)$&
  \\
   \end{tabular*}
 
 \begin{tabular*}{0.9\textwidth}{@{\extracolsep{\fill} }l}
  
  $\alpha_{10}: \alpha_1, \alpha_2\Rightarrow BeRevokedDrivingLicense(PS1)$\\
  $\alpha_{11}: \alpha_1, \alpha_2\Rightarrow TakeCriminalResposibility(PS1)$\\
  $\alpha_{12}: \alpha_4\Rightarrow  TakeCriminalResposibility(PS1)$\\
  $\alpha_{13}: \alpha_4, \alpha_6\Rightarrow AggravatedPunishment(PS1)$\\
  $\alpha_{14}: \alpha_1, \alpha_2, \alpha_4\Rightarrow AggravatedPunishment(PS1)$\\
  $\alpha_{15}: \alpha_3, \alpha_{5}\Rightarrow transferToSafePlace(PS1, Injury1)$\\
  $\alpha_{16}: \alpha_3, \alpha_5, \alpha_{7}\Rightarrow doNecessaryAid(PS1, Injury1)$\\
  $\alpha_{17}: \alpha_{8}\rightarrow \neg Intoxicated(PS1)$\\
  $\alpha_{18}: \alpha_2\rightarrow \neg Sober(PS1)$\\
  $\alpha_{19}: \alpha_{15}\rightarrow LeaveCar(PS1)$\\
  $\alpha_{20}: \alpha_{9}\rightarrow \neg transferToSafePlace(PS1, Injury1)$\\
  $\alpha_{21}: \alpha_{16}\rightarrow LeaveCar(PS1)$\\
  $\alpha_{22}: \alpha_{9}\rightarrow \neg doNecessaryAid(PS1, Injury1)$\\
\end{tabular*}

Then according to Definition \ref{def-att}, we can get the following conflict relation:

$att=\{
(\alpha_{17}, \alpha_2), (\alpha_{17}, \alpha_9), (\alpha_{17}, \alpha_{10}), (\alpha_{17}, \alpha_{11}), (\alpha_{17}, \alpha_{14}),  (\alpha_{17}, \alpha_{18}), (\alpha_{17}, \alpha_{20}), \\(\alpha_{17}, \alpha_{22}), 
(\alpha_{18}, \alpha_8), (\alpha_{18}, \alpha_{17}), 
(\alpha_{19}, \alpha_9), (\alpha_{19}, \alpha_{20}), (\alpha_{19}, \alpha_{22}), 
(\alpha_{20}, \alpha_{15}), (\alpha_{20}, \alpha_{19}),\\ 
(\alpha_{21}, \alpha_9), (\alpha_{21}, \alpha_{20}), (\alpha_{21}, \alpha_{22}), 
(\alpha_{22}, \alpha_{16}), (\alpha_{22}, \alpha_{21})\}$.

\end{example}

\subsection{Legal compliance detection}
When engineers complete a whole design draft, they could use the consistency checking function to check whether this design is fully compliant with given laws and where conflicts are.  

\begin{definition}[Consistency Checking]\label{def-csscheck}
Let $(\mathcal{L}, \mathcal{K}^A, \mathcal{R}^T, n, \mathcal{P}, prin)$ be an $\LAT^\varDelta$ \\according to legal ontology $\varDelta=(T,A)$, and $\mathcal{A}$ the set of all the arguments constructed based on it. 
The ABox of $\varDelta$ is consistent w.r.t. the TBox of $\varDelta$ iff $\mathcal{A}$ is conflict-free based on attack relations, i.e., $\nexists \alpha, \beta\in\mathcal{A}$ such that $\alpha$ attacks $\beta$.
\end{definition}

If a design is completely consistent after reasoning, it means it is fully compliant with given laws. Otherwise it is not. And by tracing where there are conflicting arguments, we could know which part of the design needs modification.  

According to our running example, the $\LAT^\varDelta$ in Example \ref{exp-lesac} is not conflict-free based on attack relation, therefore the legal ontology it is based on is inconsistent. 

\subsection{Feedback for single change}

If the AV engineers wants to remain the main design of an AV and only do some minimal changes. They can find out what further legal consequences will occur with these new details by instance checking in the $\LAT^\varDelta$, and the basic to realize instance checking is the evaluation of assertions as below.  

According to an $\LAT^\varDelta$, assertions are the conclusions of arguments. 
So based on the extension of arguments, we can decide whether an assertion is accepted. Therefore, the acceptance of assertions is defined as follows. 

\begin{definition}[Assertion Acceptance]\label{def-acc-ass}
Let $(\mathcal{L}, \mathcal{K}^A, \mathcal{R}^T, n, \mathcal{P}, prin)$ be an $\LAT^\varDelta$ and $\mathcal{A}$ the set of all the arguments constructed based on it. 
An assertion $X$ is sceptically/credulously accepted under certain argumentation semantics $S$, iff $\exists A\in\mathcal{A}$, such that $A$ is sceptically/credulously justified w.r.t. $\mathcal{E}_S$ and $\conc(A)=X$.
\end{definition}

To determine whether a certain modification is consistent with current design and given laws, we translate this problem into whether a legal assertion about this AV can be accepted as a conclusion of an accepted/justified argument.

\begin{example}[Example \ref{exp-arg} continued.]\label{exp-acc-as}
Based on the attack relation among arguments, since $\lN(\alpha_2)=\emptyset$, $\lp(\alpha_8)=\{r_1\}$, 
$\lN(\alpha_9)=\{r_2\}$, \\
$\lN(\alpha_{10})=\{r_3\}$, 
$\lN(\alpha_{11})=\{r_4\}$, 
$\lN(\alpha_{14})=\{r_7\}$, \\
$\lN(\alpha_{15})=\{r_8\}$,  $\lN(\alpha_{16})=\{r_9\}$, 
$\lN(\alpha_{17})=\{r_1\}$, \\
$\lN(\alpha_{18})=\emptyset$, 
$\lN(\alpha_{19})=\{r_8\}$, 
$\lN(\alpha_{20})=\{r_2\}$, \\
$\lN(\alpha_{21})=\{r_9\}$, 
$\lN(\alpha_{22})=\{r_2\}$
, according to Definition \ref{def-order}: 

assume that based on $\leqslant$ on $\mathcal{P}$, $p_2< p_1$, then $\mathcal{D}=\{ (\alpha_{17}, \alpha_9), (\alpha_{17}, \alpha_{10}), (\alpha_{17}, \alpha_{11}),\\ (\alpha_{17}, \alpha_{14}),  (\alpha_{17}, \alpha_{20}), (\alpha_{17}, \alpha_{22}), (\alpha_{18}, \alpha_8), (\alpha_{18}, \alpha_{17}), (\alpha_{19}, \alpha_9),  (\alpha_{19}, \alpha_{20}), (\alpha_{19}, \alpha_{22}),\\ (\alpha_{21}, \alpha_9), (\alpha_{21}, \alpha_{20}), (\alpha_{21}, \alpha_{22})\}$; while assume that $p_1< p_2$, then  $\mathcal{D}'=\{(\alpha_{17}, \alpha_9), \\ (\alpha_{17}, \alpha_{10}), (\alpha_{17}, \alpha_{11}), (\alpha_{17}, \alpha_{14}),  (\alpha_{17}, \alpha_{20}), (\alpha_{17}, \alpha_{22}), (\alpha_{18}, \alpha_8), (\alpha_{18}, \alpha_{17}), \\(\alpha_{20}, \alpha_{15}), (\alpha_{20}, \alpha_{19}), (\alpha_{22}, \alpha_{16}), (\alpha_{22}, \alpha_{21})\}$. 

According to Definition \ref{def-semantics}, for $AF=(\mathcal{A}, \mathcal{D})$, we can get  \\$\mathcal{E}_{co/pr/gr}=\{E=\{\alpha_{1-7}, \alpha_{10-16}, \alpha_{18}, \alpha_{19}, \alpha_{21}\}\}$; while for $AF'=(\mathcal{A}, \mathcal{D}')$, we can get  $\mathcal{E}_{co/pr/gr}=\{E'=\{\alpha_{1-7}, \alpha_{9-14}, \alpha_{18}, \alpha_{20}, \alpha_{22}\}\}$. 
\end{example}

For instance, consider the assertion ``$LeaveCar(PS1)$'', if $p_2< p_1$ based on $\leqslant$ on $\mathcal{P}$, then after argument evaluation, we can find a sceptically justified argument $\alpha_{19}$ (or $\alpha_{21}$) such that $\conc(\alpha_{19})=LeaveCar(PS1)$ under all the basic argumentation semantics introduced in Example \ref{def-semantics}. On the contrary, if $p_1< p_2$ based on $\leqslant$ on $\mathcal{P}$, then both $\alpha_{19}$ and $\alpha_{21}$ that conclude ``$LeaveCar(PS1)$'' are rejected. Instead, there is an argument $\alpha_{9}$ which concludes ``$\neg LeaveCar(PS1)$'' is sceptically justified. 

The following definition defines instance checking based on an $\LAT^\varDelta$ for all the possible forms of classes.

\begin{definition}[Instances Checking]
\label{def-insck-C}
Let $\varphi$ be an individual, sceptically or credulously, it holds that $\varphi$ is an instance of class:
\begin{itemize}
    \item $C$ (/$\neg C$), iff $\exists \alpha\in\mathcal{A}$, s.t. $\alpha$ is sceptically/credulously justified w.r.t. $\mathcal{E}_S$ and \\$\conc(A)=C(\varphi) (/\neg C(\varphi)$;
    \item $C\sqcap D$, iff $\exists \alpha, \beta\in\mathcal{A}$ such that $\alpha$ and $\beta$ are both sceptically/credulously justified w.r.t. $\mathcal{E}_S$ and $\conc(A)=C(\varphi)$, $\conc(B)=D(\varphi)$;
    \item $C\sqcup D$, iff $\exists \alpha, \beta\in\mathcal{A}$ s.t. at least one of $\alpha$ and $\beta$ are sceptically/credulously justified w.r.t. $\mathcal{E}_S$ and $\conc(\alpha)=C(\varphi)$, $\conc(\beta)=D(\varphi)$;
    \item $\exists P.D$, iff $\exists \alpha, \beta\in\mathcal{A}$ such that $\alpha$ and $\beta$ are both sceptically/credulously justified w.r.t. $\mathcal{E}_S$ and $\conc(\alpha)=P(\varphi,x)$, $\conc(\beta)=D(x)$ ($x$ is an individual);
    \item $\forall P.D$, iff 1) $\exists \alpha\in\mathcal{A}$ s.t. $\conc(\alpha)=P(\varphi,x)$; and 2) $\forall\alpha\in\{\alpha|\conc(\alpha)=P(\varphi,x)\}$, $\exists \beta\in\mathcal{\alpha}$, s.t. $\conc(\beta)=D(x)$.
\end{itemize}
\end{definition}

\subsection{Extract consistent design}
In some situations, facing an inconsistent design draft, engineers may want to get a set of design elements that can be collectively accepted rather than an answer for a certain query. If for instance we do not know if the AV as currently designed respects the traffic lights, we can find out how this vehicle would fare with different developments. This task can be achieved through the argument evaluation procedure based on \lasp. Extensions of arguments can be obtained according to different standards, i.e. argumentation semantics. 

\begin{definition}[Collective acceptance]\label{def-coacc}
Let $\mathcal{E}_S$ be a set of extensions under semantic $S$, for all $E_S\in\mathcal{E}_S$, $O_S=\{\conc(\alpha)|\alpha\in E_S\}$ is a set of collectively accepted design elements.

\end{definition}
 
 Consider Example \ref{exp-acc-as}, for an $AF=(\mathcal{A}, \mathcal{D})$, $E=\{\alpha_{1-7}, \alpha_{10-16}, \alpha_{18}, \alpha_{19}, \alpha_{21}\}$ is the only complete extension, and the corresponding conclusion is \\$\mathcal{K}^A\cup\{
 BeRevokedDrivingLicense(PS1), TakeCriminalResposibility(PS1),   \\AggravatedPunishment(PS1), transferToSafePlace(PS1, Injury1), \\doNecessaryAid(PS1, Injury1), \neg Sober(PS1), LeaveCar(PS1)\}$, \\which is a collectively acceptable set of design elements. 
 
 In legal contexts, sceptical reasoning is usually more appropriate for justifying beliefs (or facts), whereas for normative reasoning the users would expect to get some alternatives, for which credulous reasoning may be more appropriate. 
 According to Definition \ref{def-semantics}, among all argumentative semantics based on complete extensions, in order to get sceptical results, one can choose to apply the grounded semantics, while in order to get credulous results, the preferred semantics may be more suitable. 
 
\subsection{Manage multiple design drafts}
Consider that in some cases, engineers want to obtain all design drafts that are instances of a concept $ C $, such as all drafts where the AV assumes criminal responsibility, or want to know all the features that a design $x$ has, the following two methods can be applied to answer the above two questions respectively.

Given an $\LAT^\varDelta$, we can get a set $\mathcal{E}_S$ of all the extensions under argumentation semantic $S$. For an extension $E_S\in\mathcal{E}_S$, let $O_{E_S}=\{\conc(\alpha)|\alpha\in E_S\}$ be the set of conclusions according to $E_S$, we have

\textit{Method 1}: to obtain all the individuals that are instances of concept $C$, we find all the individual $x$ such that $C(x)\in O_{E_S}$ based on Definition \ref{def-acc-ass}.

\textit{Method 2}: to obtain all the concepts that $x$ is an instance of, we find all the concepts $C$ such that $C(x)\in O_{E_S}$ based on Definition \ref{def-insck-C}.

\subsection{Giving legal explanations}
How to best use argumentation theory to generate understandable explanations has become an increasingly important topic in AI regulation and AI design. There have been works \cite{DC18,expla} discussing which standards should an argumentation-based explanation meet and some formal explanations \cite{expla,garcia14,odstnCLAR20} for different argumentation frameworks are given. 
For legal systems, users may need explanations with understandable legal information for the acceptance of certain assertions, rather than the acceptance of a set of design elements. 
Considering the needs of AV engineers, the explanation of the reasoning results based on an $\LAT^\varDelta$ should show how a legal conclusion is obtained, and which contents in this situation make it accepted or not accepted. So we give a formal definition of explanation as follows, which includes related knowledge and legal rules for obtaining this result, as well as other crucial information like the associated legal principles that determine preferences among arguments.

For any agent or user $y$, we use $\leqslant_y$ to denote $y$'s priority orderings on the set $\mathcal{P}$. 
To explain the acceptance of an assertion, we give an explanation of why an argument $\alpha$ that concludes this assertion is accepted.

\begin{definition}[Explanation]\label{def-explanation}
Let $X$ be an assertion in an $\LAT^\varDelta$ that is \\sceptically/credulously accepted under certain argumentation semantics $S$ by a rational agent $y$, then there $\exists \alpha\in \mathcal{A}$ such that  $\conc(\alpha)=X$. The \textit{explanation} for $y$ to accept $X$ is $Exp_y=\mathcal{C}(\alpha)\cup\mathcal{C}(\beta)\cup\{\leqslant_y\}$, where:
\begin{itemize}
    \item $\mathcal{C}(\alpha)=\prem(\alpha)\cup \rules(\alpha)$, which explains how $X$ is reached;
    \item $\mathcal{C}(\beta)=\prem(\beta)\cup \rules(\beta)$ such that 
    $\beta$ defends $\alpha$ according to $\leqslant_y$ and the defeat relation $\mathcal{D}$, which explains why $X$ is justified.
\end{itemize}
\end{definition}

Definition \ref{def-explanation} provides a formal explanation of why a legal conclusion $X$ is accepted for certain design requirement. It consists of two parts. The first part explains how $X$ is reached by presenting all the premises contained in $\mathcal{K}^A$ and all the legal rules contained in $\mathcal{R}^T$ which are applied in the construction of argument $\alpha$. In other words, it reveals all the declarations contained in the original legal ontology that are used to conclude $X$. The second part explains why this legal conclusion is accepted by presenting all the legal information and relevant legal principles applied to construct arguments that defend $\alpha$. Similarly, this explanation reveals all the relevant declarations of the content from the original legal ontology.

Consider our running examples, for the acceptance of assertion ``$LeaveCar(PS1)$'', $Exp_y=(\{Injury(Injury1), CauseAccident(PS1), NeedEmergencyAid(Injury1)\}\cup\\\{r_8, r_9\})\cup\{p_2< p_1\}$; and for the acceptance assertion ``$\neg LeaveCar(PS1)$'', \\$Exp'_y=(\{Intoxicated(PS1)\}\cup\{r_2\})\cup(\{Intoxicated(PS1)\}\cup\{r_{10}'\})\cup\{p_1< p_2\}$.

\section{Conclusion}
In the current paper we mainly consider dealing with the problem of reasoning based on inconsistent ontologies, with a particular focus on reasoning with legal ontologies. We streamline the \asp framework \cite{prakken10,MP13} for the needs of normative reasoning in legal contexts, while also extending it by associating the argumentation system with a set of legal principles. Based on the associated legal principles, we can prioritize relevant legal norms and get preferences between the constructed arguments. Then the present paper merges this argumentation framework (\lasp) with the DL-based legal ontology, resulting in a combined argumentation theory, and defines some basic reasoning functions. In addition, we give a definition of explanation for why certain assertions are justifiable based on the argumentative theory. Throughout the paper, we explain our ideas and illustrate the relevant formal definitions and reasoning functions by a case study on the design of autonomous vehicles based on traffic regulations aimed at providing legal guidance to the engineers.

\bibliographystyle{splncs04}
\bibliography{dlbib}

\begin{thebibliography}{10}
\providecommand{\url}[1]{\texttt{#1}}
\providecommand{\urlprefix}{URL }
\providecommand{\doi}[1]{https://doi.org/#1}

\bibitem{dlhandbook2003}
Baader, F., Calvanese, D., McGuinness, D.L., Nardi, D., Patel-Schneider, P.F.
  (eds.): The description logic handbook: Theory, implementation, and
  applications. Cambridge University Press (2003)

\bibitem{Breuker2005}
Breuker, J., Valente, A., Winkels, R.: Use and reuse of legal ontologies in
  knowledge engineering and information management. In: Benjamins, V.R.,
  Casanovas, P., Breuker, J., Gangemi, A. (eds.) Law and the Semantic Web:
  Legal Ontologies, Methodologies, Legal Information Retrieval, and
  Applications. pp. 36--64. Springer Berlin Heidelberg, Berlin, Heidelberg
  (2005)

\bibitem{Cayrol92}
Cayrol, C., Royer, V., Saurel, C.: Management of preferences in
  assumption-based reasoning. In: Bouchon-Meunier, B., Valverde, L., Yager,
  R.R. (eds.) IPMU '92---Advanced Methods in Artificial Intelligence. pp.
  13--22. Springer Berlin Heidelberg, Berlin, Heidelberg (1992)

\bibitem{DC18}
Dauphin, J., Cramer, M.: Aspic-end: Structured argumentation with explanations
  and natural deduction. In Elizabeth Black, Sanjay Modgil and Nir Oren eds.,
  Theory and Applications of Formal Argumentation. TAFA 2017. Lecture Notes in
  Computer Science, Vol. 10757. Cham: Springer International Publishing, 2018:
  51--66

\bibitem{dung95}
Dung, P.M.: On the acceptability of arguments and its fundamental role in
  nonmonotonic reasoning, logic programming and n-person games. Artificial
  Intelligence  \textbf{77}(2),  321 -- 357 (1995)

\bibitem{dungkt09}
Dung, P.M., Kowalski, R.A., Toni, F.: Assumption-based argumentation. In:
  Simari, G., Rahwan, I. (eds.) Argumentation in Artificial Intelligence. pp.
  100--218. Springer US, Boston, MA (2009)

\bibitem{expla}
Fan, X., Toni, F.: On computing explanations in argumentation. In: Proceedings
  of the Twenty-Ninth AAAI Conference on Artificial Intelligence. p.
  1496–1492. AAAI'15, AAAI Press (2015)

\bibitem{Gangemi2005}
Gangemi, A., Sagri, M.T., Tiscornia, D.: A constructive framework for legal
  ontologies. In: Benjamins, V.R., Casanovas, P., Breuker, J., Gangemi, A.
  (eds.) Law and the Semantic Web: Legal Ontologies, Methodologies, Legal
  Information Retrieval, and Applications. pp. 97--124. Springer Berlin
  Heidelberg, Berlin, Heidelberg (2005)

\bibitem{GS04}
Garc\'{\i}a, A.J., Simari, G.R.: Defeasible logic programming: An argumentative
  approach. Theory and Practice of Logic Programming  \textbf{4}(2),  95--138
  (2004)

\bibitem{garcia14}
Garc{\'i}a, A.J., Simari, G.R.: Defeasible logic programming: Delp-servers,
  contextual queries, and explanations for answers. Argument \& Computation
  \textbf{5},  63--88 (2014)

\bibitem{simari10aai}
G{\'{o}}mez, S.A., Ches{\~{n}}evar, C.I., Simari, G.R.: Reasoning with
  inconsistent ontologies through argumentation. Applied Artificial
  Intelligence  \textbf{24}(1{\&}2),  102--148 (2010)

\bibitem{grosof03}
Grosof, B.N., Horrocks, I., Volz, R., Decker, S.: Description logic programs:
  Combining logic programs with description logic. In: Proceedings of the 12th
  International Conference on World Wide Web. p. 48–57. Association for
  Computing Machinery, New York, NY, USA (2003)

\bibitem{Hoekstra07}
Hoekstra, R., Breuker, J., Bello, M.D., Boer, A.: The {LKIF} core ontology of
  basic legal concepts. In: Casanovas, P., Biasiotti, M.A., Francesconi, E.,
  Sagri, M. (eds.) Proceedings of the 2nd Workshop on Legal Ontologies and
  Artificial Intelligence Techniques. {CEUR} Workshop Proceedings, vol.~321,
  pp. 43--63. CEUR-WS.org (2007)

\bibitem{Hoekstra09}
Hoekstra, R., Breuker, J., Bello, M.D., Boer, A.: {LKIF} core: Principled
  ontology development for the legal domain. In: Breuker, J., Casanovas, P.,
  Klein, M.C.A., Francesconi, E. (eds.) Law, Ontologies and the Semantic Web -
  Channelling the Legal Information Flood. vol.~188, pp. 21--52. {IOS} Press
  (2009)

\bibitem{luyu20dl}
Lu, Y., Yu, Z.: Argumentation theory for reasoning with inconsistent ontologies
  (extended abstract). In: Borgwardt, S., Meyer, T. (eds.) Proceedings of the
  33rd International Workshop on Description Logics (DL 2020), Online,
  September 12th to 14th, 2020. {CEUR} Workshop Proceedings, vol.~2663.
  CEUR-WS.org (2020)

\bibitem{MP13}
Modgil, S., Prakken, H.: A general account of argumentation with preferences.
  Artificial Intelligence  \textbf{195},  361--397 (2013)

\bibitem{MPtutorial}
Modgil, S., Prakken, H.: The aspic+ framework for structured argumentation: a
  tutorial. Argument {\&} Computation  \textbf{5}(1),  31--62 (2014)

\bibitem{odstnCLAR20}
Oliveira, T., Dauphin, J., Satoh, K., Tsumoto, S., Novais, P.: Goal-driven
  structured argumentation for patient management in a multimorbidity setting.
  In: Dastani, M., Dong, H., van~der Torre, L. (eds.) Logic and Argumentation.
  pp. 166--183. Springer International Publishing, Cham (2020)

\bibitem{prakken10}
Prakken, H.: An abstract framework for argumentation with structured arguments.
  Argument {\&} Computation  \textbf{1}(2),  93--124 (2010)

\bibitem{prakken97}
Prakken, H., Sartor, G.: Argument-based extended logic programming with
  defeasible priorities. Journal of Applied Non-Classical Logics
  \textbf{7}(1-2),  25--75 (1997)

\bibitem{ALC91}
Schmidt-Schau\ss{}, M., Smolka, G.: Attributive concept descriptions with
  complements. Artificial Intelligence  \textbf{48}(1),  1--26 (1991)

\bibitem{Valente95}
Valente, A.: Legal Knowledge Engineering: A Modelling Approach. IOS Press
  (1995)

\bibitem{Valente99}
Valente, A., Breuker, J., Brouwer, B.: Legal modeling and automated reasoning
  with on-line. International Journal of Human-Computer Studies  \textbf{51},
  1079--1125 (1999)

\bibitem{VanEngers2008}
Van~Engers, T., Boer, A., Breuker, J., Valente, A., Winkels, R.: Ontologies in
  the legal domain. In: Digital Government: E-Government Research, Case
  Studies, and Implementation. pp. 233--261. Springer US, Boston, MA (2008)

\bibitem{ye95}
Ye, L.R., Johnson, P.E.: The impact of explanation facilities on user
  acceptance of expert systems advice. MIS Quarterly  \textbf{19}(2),  157--172
  (1995)

\end{thebibliography}

\end{document}